# Improvement of Classification in One-Stage Detector

Wu Kehe, Chen Zuge*, Zhang Xiaoliang, Li Wei

North China Electric Power University

**Abstract:** RetinaNet proposed Focal Loss for classification task and improved one-stage detectors greatly. However, there is still a gap between it and two-stage detectors. We analyze the prediction of RetinaNet and find that the misalignment of classification and localization is the main factor. Most of predicted boxes, whose IoU with ground-truth boxes are greater than 0.5, while their classification scores are lower than 0.5, which shows that the classification task still needs to be optimized. In this paper we proposed an object confidence task for this problem, and it shares features with classification task. This task uses IoUs between samples and ground-truth boxes as targets, and it only uses losses of positive samples in training, which can increase loss weight of positive samples in classification task training. Also the joint of classification score and object confidence will be used to guide NMS. Our method can not only improve classification task, but also ease misalignment of classification and localization. To evaluate the effectiveness of this method, we show our experiments on MS COCO 2017 dataset. Without whistles and bells, our method can improve AP by 0.7% and 1.0% on COCO validation dataset with ResNet50 and ResNet101 respectively at same training configs, and it can achieve 38.4% AP with two times training time. Code is at: http://github.com/chenzuge1/RetinaNet-Conf.git.

**Key Words:** RetinaNet; misalignment; object confidence; classification improvement; IoU

## 1 Introduction

As a fundamental task in Computer Vision(CV), object detection plays an important role in CV field, and it can be used in many scenes, such as fault detection, video surveillance, disease detection, etc. Compared with image classification, object detection extra adds a localization subtask, so it belongs to multi-task learning domain. Since R-CNN[1] was proposed in 2013, deep learning has been used in object detection task nearly seven years. Through the efforts of many experts and scholars over these years, the research subject has been greatly improved. The state-of-the-arts detectors can achieve more than 80% mean average precision(mAP) on Pascal VOC[2] dataset, and some excellent detectors can achieve nearly 50% AP on COCO[3] dataset. We are glad to see these remarkable achievements, but what's more, we should pay more attention to the problems among them. One of the critical problems is misalignment between classification subtask and localization subtask in object detection. As ImageNet[4] dataset includes millions of training images used for image classification task, however, due to the high labeling cost, object detection datasets, such as VOC and COCO, have much less training images and classes. Generally the backbone network of detectors is pretrained on ImageNet dataset to learn more abundant features from massive data, and classification subtask and localization subtask will share these features. But classification task is insensitive to location of object, namely, no matter where an object appears in image, it can always be recognized correctly. And localization subtask is sensitive to location of object, so it should predict different location results once object moves. In consequence, the two subtasks of object detection may be misalignment: the predicted boxes with higher classification confidence have worse IoU(Intersection over Union) with ground truth boxes, while these boxes with better IoU have lower classification confidence. Also the predicted results would be filtered by NMS(Non-Maximum Suppression) method and it selects boxes according to their classification confidence, which results in the boxes with higher classification confidence will be preserved and the boxes with better IoU are be restricted. So the localization performance of detectors would be unsatisfied[5]. However, the boxes with accurate localization and correct classification are expected.

At present, the misalignment problem is mainly discussed in two-stage object detection models, however, it also appears in one-stage object detectors. RetinaNet[6] is the state-of-the-arts of one-stage models, Figure 1 shows the distribution of classification confidence and IoU of its prediction results in COCO dataset.

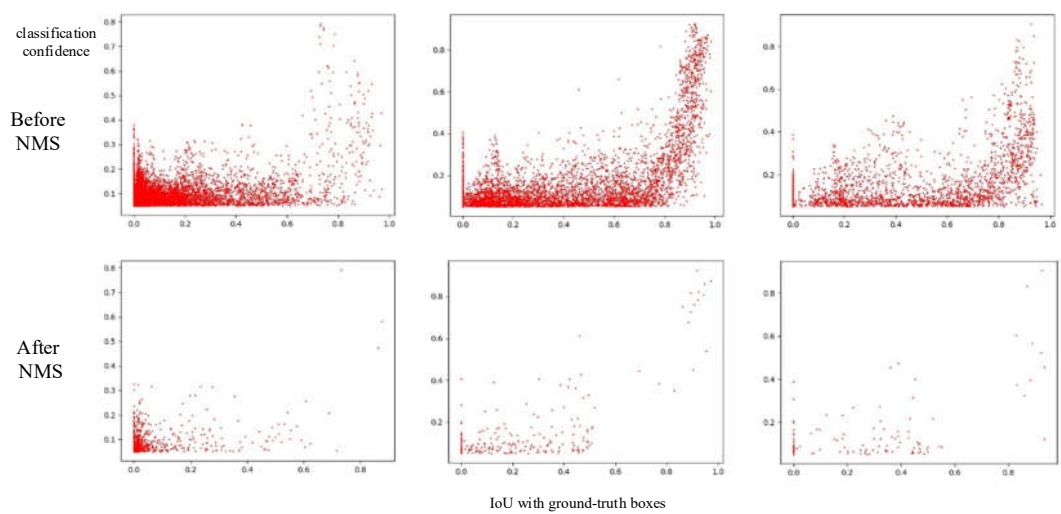

Figure 1 The distribution of classification confidence and IoU of RetinaNet's detection results in COCO dataset. The first line shows the predicted results before NMS, and their classification confidences are greater than 0.05. While the second line shows the filtered detection results by NMS, and the threshold of NMS is 0.5. And the three columns represent three different images in COCO respectively. Generally, the predictions are credible when their classification confidence are greater than 0.5.

From Figure 1 we can see that the classification confidence and IoU is also misalignment in detection results of RetinaNet. When IoU is lower than 0.7, most of the classification confidences distribute in range of 0.05~0.2, and a very small percentage of them distribute in range of 0.2~0.4 and few of them is greater than 0.4 or 0.5. When IoU is greater than 0.7, the classification confidences increase sharply. Though their IoUs are same, the classification confidences of these boxes vary from 0 to 1, also most of them are lower than 0.5. The above phenomenon demonstrates that there is no positive correlation between classification prediction and localization prediction in RetinaNet, whether the IoU is small or large. **What's more, we can see that the main factor of misalignment in RetinaNet is poor classification subtask, because the proportion of prediction results with classification confidence greater than 0.5 is far lower than that of predicted results with IoU greater than 0.5 .** The boxes with IoUs greater than 0.5 are defined as positive samples, but the prediction results show that the classification confidences of most positive boxes are lower than 0.5, which should be close to 1. Table 1 shows the quantity statistics of boxes under different conditions from ten images in COCO.

| stage | condition | Image1 | Image2 | Image3 | Image4 | Image5 | Image6 | Image7 | Image8 | Image9 | Image10 |
|---|---|---|---|---|---|---|---|---|---|---|---|
| Positive_NUM | iou>0.5 | 331 | 320 | 170 | 198 | 84 | 77 | 131 | 210 | 310 | 110 |
| Before NMS | cls>0.05 | 7213 | 2543 | 3105 | 3539 | 634 | 364 | 1700 | 13480 | 5422 | 1067 |
|  | cls>0.5 | 110 | 79 | 140 | 23 | 30 | 83 | 35 | 29 | 363 | 135 |
|  | cls>0.6 | 71 | 38 | 84 | 14 | 15 | 75 | 28 | 12 | 264 | 107 |
|  | cls>0.7 | 36 | 18 | 47 | 7 | 0 | 55 | 21 | 8 | 172 | 82 |
|  | cls>0.8 | 16 | 4 | 13 | 1 | 0 | 38 | 10 | 0 | 63 | 57 |
|  | cls>0.9 | 3 | 1 | 6 | 0 | 0 | 26 | 1 | 0 | 10 | 21 |
|  | iou>0.5 | 1087 | 1116 | 856 | 507 | 187 | 249 | 336 | 549 | 2004 | 487 |
|  | iou>0.6 | 912 | 963 | 656 | 384 | 136 | 228 | 267 | 340 | 1629 | 449 |
|  | iou>0.7 | 650 | 714 | 423 | 271 | 82 | 191 | 185 | 212 | 1197 | 372 |
|  | iou>0.8 | 307 | 473 | 238 | 167 | 52 | 152 | 98 | 111 | 816 | 208 |
|  | iou>0.9 | 105 | 136 | 23 | 49 | 14 | 77 | 17 | 28 | 324 | 68 |
| After NMS | cls>0.05 | 429 | 113 | 66 | 188 | 35 | 14 | 76 | 836 | 224 | 44 |
|  | cls>0.5 | 5 | 5 | 4 | 1 | 1 | 2 | 1 | 2 | 13 | 5 |
|  | cls>0.6 | 5 | 3 | 4 | 1 | 1 | 2 | 1 | 1 | 12 | 5 |
|  | cls>0.7 | 3 | 2 | 3 | 1 | 0 | 2 | 1 | 1 | 10 | 5 |
|  | cls>0.8 | 2 | 2 | 3 | 1 | 0 | 2 | 1 | 0 | 6 | 5 |
|  | cls>0.9 | 1 | 1 | 3 | 0 | 0 | 1 | 1 | 0 | 1 | 3 |
|  | iou>0.5 | 16 | 14 | 6 | 10 | 3 | 4 | 10 | 17 | 20 | 8 |
|  | iou>0.6 | 13 | 10 | 6 | 6 | 2 | 3 | 4 | 8 | 16 | 8 |
|  | iou>0.7 | 11 | 10 | 6 | 4 | 0 | 2 | 4 | 4 | 15 | 6 |
|  | iou>0.8 | 5 | 10 | 5 | 3 | 0 | 2 | 2 | 2 | 14 | 4 |
|  | iou>0.9 | 4 | 4 | 5 | 1 | 0 | 1 | 0 | 0 | 9 | 2 |

Table 1 The quantity statistics of boxes under different conditions from ten images in different stages of RetinaNet . The 'Positive_NUM' line shows that the number of positive default anchor boxes in each image, whose IoUs with ground-truth(gt) boxes are greater than 0.5. Classification task uses these positive samples and other negative samples as training samples and localization task only uses these positive samples as training samples. The 'Before NMS' and 'After NMS' show the number of prediction results in different inferring stage. And the statistics of all results gained by RetinaNet model directly are showed in 'Before NMS' lines, while the statistics of results filtered by NMS method are showed in 'After NMS' lines. In 'condition' column, 'iou' represents IoUs of default anchor boxes or prediction boxes with gt boxes, and 'cls' represents classification scores of prediction boxes.

In object detection models, if IoUs of default anchor boxes with ground-truth boxes are greater than 0.5, these boxes would be defined as positive samples, and the total loss of localization subtask only contains losses of these positive anchors, while the total loss of classification subtask will contain losses of these positive samples and other negative samples. From 'Positive_NUM' line in Table 1 we can see that each image has hundreds of positive anchors, while the number of total anchors in each image is almost close to 100000, such as Image9 has 76725 anchors and Image10 has 92070 anchors, which indicates there is a large margin in numbers of positive samples and negative samples. So imbalance of positive samples and negative samples is a critical problem in one-stage object detectors. 'Before NMS' lines show the numbers of boxes under different classification confidences and IoU thresholds, and we can find prediction boxes with cls>0.5 are less than predefined positive anchor boxes in most images, while prediction boxes with iou>0.5 are more than predefined positive anchors in all images. In hence, we can infer that RetinanNet can optimize location and achieve more accurate localization, but it still needs to improve classification subtask further.

The main contribution of RetinaNet is Focal Loss, it is instead of Cross Entropy loss to alleviate the imbalance between positive and negative samples in classification subtask. Focal Loss improves the performance of one-stage detectors indeed, however, we find classification performance is still unsatisfied from Table 1, which is the main factor of restricting detection performance. Classification accuracy is still not good enough so that there is misalignment between the two subtasks. From table 1 we can see that the proportion of boxes with iou>0.5 decreases fast after NMS, such as it drops by 31.5% and 39.84% in Image1 and Image6 respectively and drops by 19.52% on average in total ten images, which show that the boxes with better localization are be suppressed in NMS. In same way, we calculate variation of boxes with cls>0.5 after NMS, the proportions of Image2, Image3 and Image8 increase by 1.32%, 1.55% and 0.02% respectively, and the proportions drop in other images, and it drops by 1.09% on average in all images. The proportions of boxes under different conditions are showed in Figure 2.

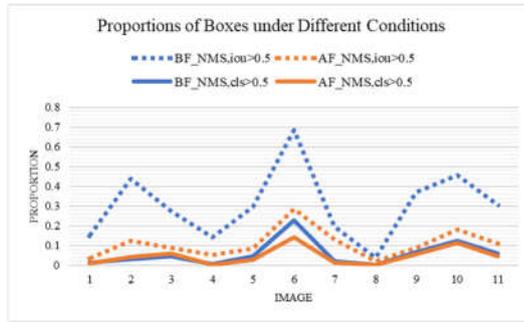

Figure 2 The proportions of boxes under different conditions. In Figure 2, 'BF_NMS' represents 'before NMS' and 'AF_NMS' represents 'after NMS'. Blue lines show proportions of boxes before NMS, and orange lines show proportions of boxes after NMS. Dotted lines show proportions of boxes filtered by IoU threshold and solid lines show proportions of boxes filtered by classification confidence threshold.

Above analyses show that NMS has more negative effect on localization performance than classification performance. The essential cause is NMS selecting boxes mainly according to their classification scores, so the boxes with higher classification scores are retained, but there is misalignment between localization and classification, which leads to the boxes with higher IoU and lower classification scores are abandoned. Therefore, only the index guiding NMS contains IoU information and it is increased might the boxes with better localization confidence be chose. Besides, Figure 2 shows that boxes with iou>0.5 are far more than the boxes with cls>0.5 before NMS, and Table 1 shows that the number of boxes with cls>0.5 before NMS is less than that of predefined positive anchor boxes, which indicates the performance of classification subtask should have a vast room for improvement. Based on this, we focus on improving classification confidence of RetinaNet to solve the misalignment between classification and localization in this paper.

In this paper we introduce a method for RetinaNet to alleviate the misalignment of two subtasks through improving classification task performance. The method adds an extra object confidence subtask in RetinaNet, and the new subtask shares features with original classification task. For convenience of expression in next introduction, the new model is called as RetinaNet-Conf, and the network architecture of it is explained directly in Section 3. To summarize, the contributions of this paper are as follows:

1) We delve into the essential causes of misalignment problem in one-stage detector RetinaNet and

reveal that poor classification performance is the critical factor.

2) We add a object confidence subtask in RetinaNet and name the new model as RetinaNet-Conf. Object confidence task shares features with classification task, which can improve the representation of classification.

3) We combine object confidence and classification score to guide NMS in inference. The combination contains information of them, so it can ease misalignment of classification and localization.

4) We validate the performance of RetinaNet-Conf on COCO dataset. It can gain 36.1% AP with ResNet-50 backbone and 38.0% AP with ResNet-101 backbone. With two times training time, it can gain 38.4% AP with ResNet-101 backbone.

## 2 Related Work

**Object Detectors:** Since deep learning is introduced from 2014, object detection has achieved abundant accomplishments. Now these detectors mainly are divided into three parts: two-stage detectors, one-stage detector and anchor free detectors. At the beginning of object detection employing deep learning, two-stage detectors are popular. As the name suggests, it contains two steps in training: 1) generating candidate boxes; 2) refining the classification and localization of these candidate boxes. Its representative detectors include RCNN[1], SPPNet[7], Fast RCNN[8], Faster RCNN[9], FPN[10]. They have more accurate localization and correct classification, at same time, they also need much more time in training and inferring. To improve real-time and availability of detectors, one-stage detectors are proposed. They drop the step of generating candidate boxes, and predict class and location of objects according to default anchor boxes directly. YOLO v1[11], YOLO V2[12], YOLO V3[13], SSD[14], RetinaNet[6] are all classical one-stage detectors. These detectors have an great improvement in executive efficiency. But most of default anchor boxes are negative samples, so the imbalance between positive and negative samples is the key factor of their unsatisfactory performance. Anchor free detectors drop default anchor boxes further, they predict results according to feature maps directly, so it needs less human experience and performs better efficiency. Its representative detectors include CornerNet[15], ExtremeNet[16], CenterNet[17], FCOS[18]. Compared with one-stage detectors, not only they can reduce computation cost, but also they can ease imbalance of training samples partly. Now anchor free detector is at the center of research, and it is worthy to be paid more attention to.

**Misalignment of Detectors:** Misalignment problem is proposed in IoU-Net firstly. It introduced that there is no positive correlation between classification and localization in FPN objectors. NMS can suppress the boxes with more accurate location, so the filtered boxes would have inaccurate location. To solve this problem, it added an IoU branch in FPN to predict localization confidence, then it would choose final boxes according to IoU prediction in NMS. Softer-NMS[19] was also concerned about the problem at same time. But different from IoU-Net, it paid more attention to the drawbacks of NMS method and proposed an evaluation method of localization according to distribution of boxes. It increased the localization confidence through minimizing variance of predicted boxes. And it used the variance to fuse multiple filtered boxes in NMS, which can improve localization further. Cascade RCNN[20] pointed out using single IoU threshold to distinguish positive and negative samples is the key constraint of localization capacity in detectors. So it proposed a multi-stage-prediction detector to address the problem. Each stage has own IoU threshold, and the IoU threshold increases as stage is deeper. The detector can improve localization confidence of prediction results, in hence, it provided a solution for misalignment problem at a different view point. Recently, more and more researchers pay attention to this subject. Yue Wu et al.[21] considered that classification and localization subtasks use same head structures, such as fully connected head(fc-head) or convolution head(conv-head), in RCNN based detectors. However, they observed that fc-head is more suitable for classification task and conv-head is more suitable for localization task, so they used different heads for specific tasks. The experiment results showed the validation of their idea. Inspired by double-head RCNN, Guanglu Song et al.[22] delved into the misalignment problem in RCNN based detectors. They illustrated that classification task is more sensitive to features of salient area while localization task is more relevant to boundary features. So they proposed a task-aware spatial disentanglement(TSD) operator, which is used after RoI(region of interesting) pooling in Faster RCNN. The operator can generate specific features for different tasks from RoI pooling features. Also each task added different heads after specific feature maps. The main contribution of it is that it demonstrated that the concerned features of the two subtask is different, so TSD operator generating specific features for specific task can improve the two tasks simultaneously, which can solve the misalignment problem in RoI pooling based detectors. However, the above solutions of misalignment mainly concern on two-stage detectors, there is few researches on misalignment in one-stage detector. Now we only look up a study published by Shengkai Wu et al.[23] Similar to IoU-Net, they also added a IoU prediction branch in RetinaNet, and it shared features with origin localization task, but the difference is that they use multiply of classification score and IoU prediction instead of IoU prediction to guide NMS.

From above discussions and analyses we can see that these solutions mainly focus on improving localization or predicting localization confidence, which is used to guide NMS to avoid the boxes with more accurate location being suppressed. Or the double-head RCNN and TSD redesign architectures of selecting feature for each task specifically, which solve the misalignment problem caused by feature sharing in classification and localization tasks. However, we can find that RetinanNet, which is a one-stage detector, has decoupled features for each task, and the features used for tasks are generated by its own four convolution layers after fusion features. What's more, from Figure 1 we can know that poor classification confidence is the key factor of misalignment in RetinaNet. In hence, in this paper we will propose a solution for misalignment in RetinaNet through improving classification.

## 3 Method

This section includes three parts: we introduce the architecture of RetinaNet_Conf detector in section 3.1 firstly. And section 3.2 will introduce the training of detector, which mainly includes descriptions of loss function and training samples in new object confidence subtask. Lastly, section 3.3 will mainly introduce how to use object confidence is used in inference.

### 3.1 Architecture of RetinaNet-Conf Detector

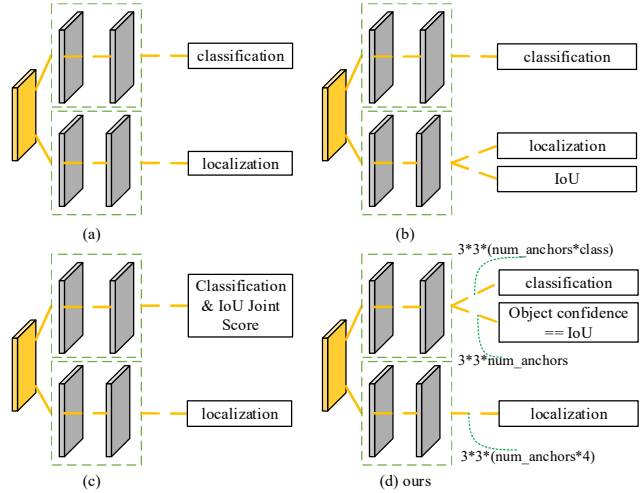

Figure 3 Comparisons of different subtask heads based on RetinaNet. (a): The initial subtask heads of RetinaNet. It only includes classification and localization subtasks, also the two tasks use different features. (b): The subtask heads are proposed by Shengkai Wu et al.[23] It adds an extra IoU subtask, which shares features with localization task. (c): Xiang Li et al.[24] proposes a special subtask instead of initial classification task, and the localization task is same with RetinaNet. (d): The subtask heads are proposed by our work, we add an extra object confidence branch, which shares features with classification task. The target of new task is same with that of IoU task proposed in (b), and it can not only improve classification but also ease misalignment through by being used in NMS.

Compared with RetinaNet, RetinaNet-Conf mainly add an object confidence branch as showed in Figure 3(d). Compared with IoU-aware Net shown in Figure 3(b), the IoU(object confidence) prediction branch shares features with classification task instead of localization task. Although they have little difference in network architecture, they play totally different roles. IoU branch in IoU-aware Net learns representation of localization confidence. When the product of IoU and classification is used to guide NMS, it can contribute to increasing the probability of the boxes with better location being selected, so as to improve localization level of predicted results. However, the product of classification and IoU used in IoU-aware Net, which is calculated by function(1) and whose function curve is shown in Figure 4, can't solve the misalignment of RetinaNet fundamentally. From Figure 4 we can see that, only when IoU and classification are large at the same time, the product would be large. Once one of them is enough small, the product would be restricted to be very low. As described in Section 1, the main cause for misalignment in RetinaNet is the inferior classification performance, besides, the new IoU branch makes few contribution to classification task, therefore classification performance is still unsatisfied.

$$product = (IoU)^{0.5} * (classification)^{0.5}$$
$$s.t. \quad 0 \le IoU \le 1, \quad 0 \le classification \le 1 \tag{1}$$

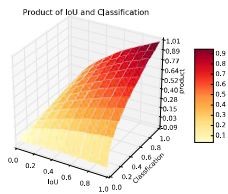

Figure 4 the product of IoU and classification

Function(2) shows that when $IoU \geq classification$ the product is greater than or equal to classification, but the IoU and classification tasks are separated in training, there is no clear relationship between them, so the improving detection performance ability of the product is limited.

$$\begin{aligned} product &\geq classification \\ (IoU * classification)^{0.5} &\geq classification \\ (IoU * classification)^{0.5} &\geq (classification * classification)^{0.5} \\ IoU &\geq classification \end{aligned} \quad (2)$$

As showed in Function(3), we can see that the maximization of the product is $(classification)^{0.5}$, when classification has no change. Compared with RetinaNet, IoU-aware Net almost has same classification, so the maximization of the product is limited by classification score.

$$\begin{aligned} product &= (IoU)^{0.5} * (classification)^{0.5} \\ &= (IoU * classification)^{0.5} \\ &\leq (1 * classification)^{0.5} \\ &\leq (classification)^{0.5} \end{aligned} \quad (3)$$

**3.2 Training**

As an extra object confidence task added, the loss of RetinaNet-Conf includes three parts: classification loss, localization loss and object confidence loss. The classification loss is the focal loss proposed in RetinaNet. Because our experiments are based on MMDetection[25], localization loss uses L1 loss instead of smooth L1 loss used in RetinaNet. The two losses are showed in function(4-5).

$$L_{classification} = -\frac{1}{N_{pos}} \cdot \sum_{i}^{N} \alpha \cdot (1-p_i)^{\gamma} \cdot \log(p_i) \quad (4)$$

$$L_{localization} = \frac{1}{N_{pos}} \cdot \sum_{i \in pos}^{N_{pos}} \sum_{m \in x,y,w,h} |y_i^{m'} - y_i^m| \quad (5)$$

In function(4-5), $N$ is total number of positive and negative samples, $N_{pos}$ is number of positive samples. $p_i$ is sigmoid value of classification prediction. $y_i^{m'}$ is prediction of location offset and $y_i^m$ is target of location offset. As same with RetinaNet, $\alpha$ is 0.25 and $\gamma$ is 2.

Object confidence loss uses cross entropy loss in our work, which is same with the loss of IoU in IoU-aware Net, next we will explain why we don't use other losses from the viewpoint of mathematical. As we all know, the IoU between two boxes ranges from 0.0 to 1.0, which is the target of object confidence task, so we need add an sigmoid function to transform output into [0.0, 1.0]. The following functions show gradients of parameter $\theta_i$ when using three different loss functions.

$$\begin{aligned} \nabla_{\theta_i}(l1\_loss) &= \frac{\partial |y - h_\theta(x)|}{\partial \theta_i} \\ &= \begin{cases} \dfrac{\partial(-h_\theta(x))}{\partial \theta_i}, & y \geq h_\theta(x) \\ \dfrac{\partial h_\theta(x)}{\partial \theta_i}, & y \leq h_\theta(x) \end{cases} \\ &= \begin{cases} -h_\theta'(x) \cdot x_i, & y \geq h_\theta(x) \\ h_\theta'(x) \cdot x_i, & y \leq h_\theta(x) \end{cases} \end{aligned} \quad (6)$$

$$\nabla_{\theta_i}(l2\_loss) = \frac{1}{2} \cdot \frac{\partial(y - h_\theta(x))^2}{\partial \theta_i} \qquad (7)$$
$$= (y - h_\theta(x)) \cdot h_\theta'(x) \cdot x_i$$

$$\nabla_{\theta_i}(cross\_entropy\_loss) = -\frac{\partial(y \cdot \log(h_\theta(x)) + (1-y) \cdot \log(1 - h_\theta(x)))}{\partial \theta_i}$$
$$= -(\frac{y}{h_\theta(x)} \cdot h_\theta'(x) \cdot x_i - \frac{1-y}{1-h_\theta(x)} \cdot h_\theta'(x) \cdot x_i)$$
$$= -(\frac{y}{h_\theta(x)} \cdot (h_\theta(x) \cdot (1 - h_\theta(x))) - \frac{1-y}{1-h_\theta(x)} \cdot (h_\theta(x) \cdot (1 - h_\theta(x)))) \cdot x_i \qquad (8)$$
$$= -(y \cdot (1 - h_\theta(x)) - (1-y) \cdot h_\theta(x)) \cdot x_i$$
$$= -(y - h_\theta(x)) \cdot x_i$$

In Function(6-8), $y$ is target of sample $x$, $h_\theta(x)$ is prediction calculated by sigmoid function. From Function(6-7) we can see that the gradients of l1 loss and l2 loss are mainly decided by $h_\theta'(x)$. With l1 or l2 loss, if $y$ is 0 and $h_\theta(x)$ is close to 1.0 or $y$ is 1 and $h_\theta(x)$ is close to 0, it shows that there is a large gap between target and prediction, so the loss should use a larger gradient to optimize network faster. However, from Figure 5 we can find the closer to 1.0 or 0 the $h_\theta(x)$ is, the smaller the gradient is. In hence, the gradient with large loss may be smaller, which would deteriorate the convergence speed. Besides, with l1 loss, when $y$ and $h_\theta(x)$ are all in range of 0.2~0.8, the target and prediction are almost same, but the gradient would be large as shown in Figure 5, so it will bring big update step for parameters, which would result in missing optimal solution.

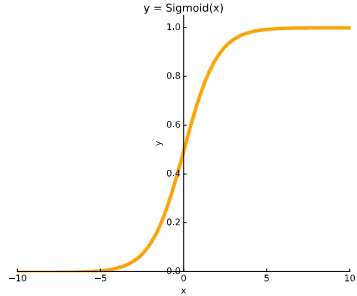

Figure 5 Sigmoid function curve

Function(8) shows the gradient calculated by cross entropy loss. We can see that the gradient mainly decided by the distance of the target and prediction. The bigger the distance is, the larger the gradient is. The bigger distance indicates there still exists a large optimization space, and the larger gradient would bring a large update step for parameters. And the smaller distances shows the prediction has been close to target, so the network only calculates a slight update by smaller gradient. So cross entropy loss is more suitable for object confidence task than l1 loss and l2 loss. Besides, as target of object confidence is the IoU of positive prediction boxes and gt boxes, so the loss of object confidence also only includes losses of positive samples. It will increase the weight of positive samples in classification head training, so as to ease imbalance of positive samples and negative samples partly. The loss function of object confidence is showed in Function(9).

$$L_{object\_confidence} = -\frac{1}{N_{pos}} \cdot \sum_{i \in pos}^{N_{pos}} (y \cdot \log(p_i) + (1-y) \cdot \log(1 - p_i)) \qquad (9)$$

In function(9), $y$ is target of object confidence, $p_i$ is sigmoid value of prediction output.

The total loss of RetinaNet-Conf can be calculated as follows:

$$L = L_{classification} + L_{localization} + L_{object\_confidence} \quad (10)$$

In practical experiments we find three losses are in the same order of magnitude, so the weights of them are all set to 1.0 in this paper.

**3.3 Inference**

As introduced above, an object confidence task is added in RetinaNet. Object confidence represents the probability of an boxes being a real object, and classification score represents the probability of an boxes being a class. The combination of them can reflect whether a box is really a class more credibly, also it contains confidence information of object and location simultaneously. As there is a sigmoid function after classification and object confidence tasks, the predictions of them are all smaller than 1.0. So the direct product of them will be smaller than either of them, if it is used to guide NMS, detection performance may be damaged. In order to there is same magnitude between product and factors, we cite the method proposed in IoU-aware Net to calculate the product of classification score and object confidence score, and it is showed in function(11). The product will be used in NMS, instead of classification score. The object confidence of positive anchor box is larger than 0.5 generally, so the product can improve the boxes with classification lower than 0.5 and IoU larger than 0.5, which promote more boxes with higher IoU to be chose in NMS and improve the object confidences of them meanwhile.

$$product = (object\_confidence)^{\alpha}(classification)^{1-\alpha}$$
$$s.t. \quad 0 \leq \alpha \leq 1 \quad (11)$$

## 4 Experiments

We present experimental results on the open COCO 2017 dataset. There are total 80 object classes in the dataset. The COCO train2017 dataset, including 118k images, is used for training and the COCO val2017 dataset, including 5k images, is used for validation. The AP and Recall metrics are used to evaluate methods in experiments. All of experiments are based on MMDetection benchmark, an excellent open source libraries of CV(computer vision). We use one GPU of TITAN RTX with 24G memory to run experiments. The input scale is [1000,600], batch size is 16 and total epochs is 12. The optimizer is SGD with 0.9 momentum and 0.0001 weight decay. The initial lr(learning rate) is 1e-2 and it will decay 0.1 times at [9, 12] epochs, and the warm-up strategy is used to adjust lr before 500 iterations.

| model | AP | $AP_{50}$ | $AP_{75}$ | $AP_S$ | $AP_M$ | $AP_L$ | AR | $AR_S$ | $AR_M$ | $AR_L$ |
|---|---|---|---|---|---|---|---|---|---|---|
| RetinaNet[6] | 0.353 | **0.543** | 0.375 | **0.184** | 0.394 | 0.480 | 0.522 | **0.308** | 0.573 | 0.694 |
| IoU-aware Net[23] | 0.355 | 0.528 | **0.379** | 0.182 | 0.399 | 0.480 | 0.525 | 0.297 | 0.578 | 0.704 |
| RetinaNet-Conf (ours) | **0.357** | 0.529 | 0.377 | 0.176 | **0.402** | **0.485** | **0.526** | 0.292 | **0.580** | **0.709** |

Table 1: Comparison of three models with ResNet50. These experiments only use classification score to guide NMS. It is used to show the effects of the extra added IoU task and object confidence task in RetinaNet. All of experiment results are retrained based on RetinaNet from MMDetection.

From Table 1 we can see that IoU-aware Net and RetinaNet-Conf all perform better than initial RetinaNet. IoU-aware Net and RetinaNet-Conf improve AP by 0.2 and 0.4 point respectively. It indicates that the IoU prediction task can always improve detection performance, no matter it share features with classification task or localization task. Compared IoU-aware Net, our work improves AP by 0.2 point, it shows the joint of IoU prediction and classification can study better representation of network. But when AP gets better, $AP_{50}$ drops by ~1.5 point and $AP_{75}$ increases slightly in either of two models. The phenomenon fully explains that the boxes filtered by NMS have higher IoU in improved detectors. Namely, more boxes with more accurate localization are chosen, so the classification scores of them are increased, which demonstrates the IoU prediction task is valid to ease misalignment of classification and localization. Also we can find AP and AR of small scale objects decline in two improved models, however, those of medium and large scale objects increase in two new models, especially in RetinaNet-Conf. The $AP_S$ declines by 0.8 and the average of $AP_M$ and $AP_L$ increases by 0.65 in RetinaNet-Conf. $AR_S$ declines by 1.6 point seriously and the average of $AP_M$ and $AP_L$ increases by 1.1 amazedly in RetinaNet-Conf. These show that the new added IoU task hurt the performance of small objects, however, it can improve the performance of medium and large objects. For small scale objects, IoU will vary great when there is a slight change in location, so it is difficult to learn suitable features when using IoU to supervise training. This is may be the main factor of its damaged performance. But for larger scale objects, IoU can be stable relatively when location changing, and the trained features are more reliable for these objects.

| loss function | AP | $AP_{0.50}$ | $AP_{0.75}$ | $AP_S$ | $AP_M$ | $AP_L$ | AR | $AR_S$ | $AR_M$ | $AR_L$ |
|---|---|---|---|---|---|---|---|---|---|---|
| L1 loss | 0.351 | 0.539 | 0.372 | 0.180 | 0.390 | 0.484 | 0.517 | 0.300 | 0.566 | 0.690 |
| Smooth L1 loss | 0.353 | **0.540** | 0.376 | 0.177 | 0.389 | 0.490 | 0.518 | 0.304 | 0.564 | 0.690 |
| L2 loss | 0.352 | 0.538 | 0.374 | 0.180 | 0.394 | 0.478 | 0.520 | 0.298 | 0.567 | 0.693 |
| GFocal loss | 0.352 | **0.540** | 0.375 | **0.188** | 0.392 | 0.472 | 0.522 | **0.307** | 0.572 | 0.689 |
| $CELoss^w$ (w=0.0001) | 0.355 | 0.529 | **0.378** | 0.179 | 0.396 | **0.488** | 0.526 | 0.290 | 0.577 | **0.710** |
| $CELoss^w$ (w=0.001) | 0.353 | 0.525 | 0.377 | 0.177 | 0.396 | 0.481 | 0.522 | 0.289 | 0.576 | 0.705 |
| CE loss | **0.357** | 0.529 | 0.377 | 0.176 | **0.402** | 0.485 | **0.526** | 0.292 | **0.580** | 0.709 |

Table 2: Comparison of different loss functions of object confidence task in RetinaNet-Conf with ResNet50. Same with Table 1, these experiments only use classification score to guide NMS. $CELoss^w$ (Cross Entropy Loss) is the cross entropy loss with an extra weight, the weight w represents the loss weight of negative samples. AS the number of negative samples is far larger than that of positive samples, so we set a very small weight for loss of negative samples to reduce the effect of them. Except for $CELoss^w$, other loss functions only compute losses of positive samples.

As introduced in RetinaNet, the focal loss function is showed in Function(12):

$$focal\_loss = \alpha \cdot (1-p)^\gamma \log(p) \qquad (12)$$

In Function(12), $p$ is probability of prediction, $\alpha$ is the loss weight of positive or negative samples, $\gamma$ is s parameter used to decay loss according to accuracy. In our experiments, we find it is difficult to converge when using negative samples to train in focal loss, so we only use positive samples in this loss function. Also the target of object confidence is continuous, not the discrete value of 0 or 1. So the GFocal loss[24] function instead of focal loss function is used in experiments, it is computed as following:

$$\begin{aligned} gfocal\_loss &= (y-p)^\gamma \cdot CELoss(p,y) \\ &= (y-p)^\gamma \cdot (y \cdot \log p + (1-y) \cdot \log(1-p)) \end{aligned} \qquad (13)$$

From Function(13) we can find that GFocal loss achieves minimum 0 when y is equal to p, so it can satisfy object confidence task.

In Table 2, the top three loss functions are generally used for computing the distance of target and prediction which are continuous values. The bottom four loss functions are CE or its transformation, and they are used for computing distribution similarity of target and prediction. The performances of CE-series are better than other three loss functions commonly, it shows that loss function used for distribution is more suitable to object confidence task. Except for $AP_{0.50}$, $AP_S$, $AR_S$, CELoss basically has the highest performance. GFocal loss has the best performance on $AP_{0.50}$, $AP_S$, $AR_S$, it shows the strategy of reducing loss weight of easy samples and increasing loss weight of hard samples plays an important role in objects with lower IoU and small scale. As introduced at above, the IoU of small object will have greater change though there is a little difference in location, so it is easy to appear bigger error in small objects, so small objects belong to hard examples. The objects with 0.5~0.75 IoU values have lower object confidences, and it is more difficult to distinguish whether it is positive sample or negative sample, so these objects also belong to hard examples. As showed in Function(13), the hard examples will have larger loss weights, so the network will learn more suitable features for these examples.

Although CELoss has the best performance, however there is some work to research. Because we only use the CELoss of positive examples in training, the supervision of negative samples is absent. The object confidence prediction of some negative samples may be larger than 0.5, which would result in wrong prediction when using product of classification and object confidence in NMS. $CELoss^w$ try to use both of positive and negative samples in training. In order to converge model, we add a loss weight w of negative samples. As shown in Table 2, the $CELoss^w$ is better when w=0.0001. Though its AP is still 0.2 smaller than CELoss, but in order to verify difference of different samples used in training, we carry out experiment in Table 3 further.

| Loss Function | AP | $AP_{0.50}$ | $AP_{0.75}$ | $AP_S$ | $AP_M$ | $AP_L$ | AR | $AR_S$ | $AR_M$ | $AR_L$ |
|---|---|---|---|---|---|---|---|---|---|---|
| $CELoss^w$ (alpha=0.4) | 0.359 | 0.528 | 0.385 | 0.186 | 0.401 | **0.494** | 0.535 | 0.310 | 0.587 | 0.714 |
| CE loss(alpha=0.4) | **0.360** | 0.529 | 0.384 | 0.184 | **0.406** | 0.491 | 0.536 | **0.316** | **0.592** | 0.714 |

Table 3: Comparison of different samples used in training of object confidence task in RetinaNet-Conf with ResNet50. The product of classification score and object confidence is used to guide NMS. $CELoss^w$ represents $CELoss^w$ (w=0.0001) in Table 2. When alpha=0.4, the performances of $CELoss^w$ and CE loss are best.

| joint | AP | $AP_{0.50}$ | $AP_{0.75}$ | $AP_S$ | $AP_M$ | $AP_L$ | AR | $AR_S$ | $AR_M$ | $AR_L$ |
|---|---|---|---|---|---|---|---|---|---|---|
| cls | 0.357 | 0.529 | 0.377 | 0.176 | 0.402 | 0.485 | 0.526 | 0.292 | 0.580 | 0.709 |
| cls(obj>0.5) | 0.358 | 0.532 | 0.379 | 0.181 | 0.402 | 0.485 | 0.531 | 0.306 | 0.586 | 0.710 |
| cls*obj | 0.359 | 0.526 | 0.384 | 0.184 | 0.406 | 0.491 | 0.535 | 0.308 | 0.593 | 0.713 |
| cls*obj(obj>0.5) | 0.359 | 0.525 | 0.383 | 0.184 | 0.407 | 0.492 | 0.534 | 0.307 | 0.593 | 0.713 |
| product (alpha=0.4,obj>0.5) | 0.360 | 0.529 | 0.384 | 0.184 | 0.406 | 0.492 | 0.535 | 0.316 | 0.593 | 0.714 |
| product(alpha=0.4) | 0.360 | 0.529 | 0.384 | 0.184 | 0.406 | 0.491 | 0.536 | 0.316 | 0.592 | 0.714 |

Table 4: Comparison of different combination methods of classification score and object confidence in Inference. "cls" represents classification score and "obj" represents object confidence score. "product" represents the joint method showed in Function(11).

"obj>0.5" is a filter condition, which used to choose the predicted boxes with greater than 0.5 object confidence.

From Table 2 and Table 3 we can see that is CELoss$^w$ and CELoss increased by 0.4 and 0.3 respectively when using product in NMS. The CELoss$^w$ promotes a little much than CELoss, and it indicates total samples can achieve better object confidence than only positive samples in training. But CELoss still performs better than CELoss$^w$ in final AP results, so we choose to use CELoss for object confidence task in next experiments. Meanwhile the more suitable loss function for object confidence, which can calculate losses of positive and negative samples, is expected to be proposed.

We can find that the product joint method performs best in the three methods in Table 3. And the direct multiple of classification and object confidence is almost same with product method with alpha. Besides, we can see that the condition of "obj>0.5" can change the results of only using classification in NMS, but it has little effect on final results, so it shows that the product method can guarantee the filtered boxes have greater than 0.5 object confidence and the product method can select more reliable detection results in NMS. Therefore the product with alpha will be used in our work.

| alpha | model | AP | AP$_{0.50}$ | AP$_{0.75}$ | AP$_S$ | AP$_M$ | AP$_L$ | AR | AR$_S$ | AR$_M$ | AR$_L$ |
|---|---|---|---|---|---|---|---|---|---|---|---|
| 0.0 | IoU-aware Net | 0.355 | 0.528 | 0.379 | 0.182 | 0.399 | 0.480 | 0.525 | 0.297 | 0.578 | 0.704 |
|  | RetinaNet-Conf | 0.357 | 0.529 | 0.377 | 0.176 | 0.402 | 0.485 | 0.526 | 0.292 | 0.580 | 0.709 |
| 0.1 | IoU-aware Net | 0.357 | 0.53 | 0.381 | 0.185 | 0.400 | 0.482 | 0.528 | 0.302 | 0.581 | 0.707 |
|  | RetinaNet-Conf | 0.359 | 0.532 | 0.380 | 0.180 | 0.403 | 0.487 | 0.531 | 0.303 | 0.583 | 0.710 |
| 0.2 | IoU-aware Net | 0.359 | 0.531 | 0.384 | 0.187 | 0.401 | 0.484 | 0.532 | 0.311 | 0.585 | 0.708 |
|  | RetinaNet-Conf | **0.360** | 0.532 | 0.382 | 0.182 | 0.404 | 0.488 | 0.533 | 0.308 | 0.587 | 0.711 |
| 0.3 | IoU-aware Net | 0.360 | 0.531 | 0.385 | 0.189 | 0.402 | 0.486 | 0534 | 0.314 | 0.589 | 0.709 |
|  | RetinaNet-Conf | **0.360** | 0.531 | 0.383 | 0.183 | 0.405 | 0.490 | 0.535 | 0.309 | 0.590 | 0.713 |
| 0.4 | IoU-aware Net | **0.361** | 0.529 | 0.388 | 0.190 | 0.404 | 0.488 | 0.536 | 0.317 | 0.591 | 0.711 |
|  | RetinaNet-Conf | **0.360** | 0.529 | 0.384 | 0.184 | 0.406 | 0.491 | 0.536 | 0.316 | 0.592 | 0.714 |
| 0.5 | IoU-aware Net | **0.361** | 0.527 | 0.389 | 0.189 | 0.406 | 0.489 | 0.538 | 0.316 | 0.593 | 0.715 |
|  | RetinaNet-Conf | 0.359 | 0.526 | 0.384 | 0.184 | 0.406 | 0.491 | 0.537 | 0.316 | 0.594 | 0.715 |
| 0.6 | IoU-aware Net | 0.359 | 0.521 | 0.390 | 0.189 | 0.406 | 0.489 | 0.538 | 0.316 | 0.593 | 0.715 |
|  | RetinaNet-Conf | 0.356 | 0.520 | 0.382 | 0.184 | 0.406 | 0.487 | 0.537 | 0.317 | 0.596 | 0.717 |
| 0.7 | IoU-aware Net | 0.354 | 0.511 | 0.385 | 0.187 | 0.407 | 0.485 | 0.538 | 0.320 | 0.596 | 0.716 |
|  | RetinaNet-Conf | 0.350 | 0.509 | 0.376 | 0.182 | 0.406 | 0.480 | 0.535 | 0.314 | 0.597 | 0.712 |
| 0.8 | IoU-aware Net | 0.340 | 0.487 | 0.371 | 0.183 | 0.403 | 0.468 | 0.531 | 0.320 | 0.592 | 0.709 |
|  | RetinaNet-Conf | 0.334 | 0.483 | 0.360 | 0.176 | 0.399 | 0.462 | 0.529 | 0.303 | 0.592 | 0.710 |
| 0.9 | IoU-aware Net | 0.283 | 0.399 | 0.311 | 0.162 | 0.381 | 0.404 | 0.498 | 0.276 | 0.559 | 0.684 |
|  | RetinaNet-Conf | 0.280 | 0.398 | 0.304 | 0.156 | 0.378 | 0.398 | 0.492 | 0.260 | 0.556 | 0.681 |

Table 5: Comparison of different alphas in product, showed in Funciton(11), of classification and object confidence in IoU-aware Net and RetinaNet-Conf with ResNet50.

Table 5 shows that RetinaNet-Conf achieves best AP when alpha is equal to 0.2, 0.3 and 0.4, while IoU-aware Net performs best at 0.4 and 0.5 alpha. It shows that there is difference in characters of object confidence and IoU tasks. In IoU-aware Net, classification and IoU are independent because they use different features, one represents class confidence and one represents location confidence, so the product perform best when they have similarly weights. But for RetinaNet-Conf, one represents class confidence and one represents object confidence in product, they influence each other because they sharing features. There is a potential relationship in them, so their product may perform best when alpha is any value. Also we find that the best AP of IoU-aware Net is 0.361, which is higher than that of RetinaNet-Conf. The cause of it is that the target of IoU task or object confidence is decided by localization task, so IoU prediction in IoU-aware Net will be more accurate than object confidence prediction because it shares features with localization task. But as described in Section 3, the product in IoU-aware Net has no practical meaning and is only a simply combination of two indexes, and there is no real change in representation of model, but the representation of classification task is improved really in our work, so our method is better than it. Table 5 also shows APs are same when alpha is equal to 0.2, 0.3 and 0.4, and except for $AP_{0.50}$, the larger is alpha, the better is other metrics, especially $AR_S$. It indicates object confidence can improve detection of boxes with higher IoU and smaller scales. Besides, Table 5 also shows AP decreases sharply with alpha increases when alpha is greater than 0.5 in two models, it expresses that classification score should be the main factor in NMS, rather than IoU. That's because many different boxes may have same IoUs, but they have total different classification score, it may mislead NMS and reduce performance of detector when using IoU as main factor in NMS.

| detector | backbone | schedule | size | AP | $AP_{0.50}$ | $AP_{0.75}$ | $AP_S$ | $AP_M$ | $AP_L$ |
|---|---|---|---|---|---|---|---|---|---|
| YOLOv2[12] | DarkNet-19 | - | [416,416] | 0.216 | 0.440 | 0.192 | 0.500 | 0.224 | 0.355 |
| YOLOv3[13] | DarkNet-53 | - | [608,608] | 0.330 | 0.579 | 0.344 | 0.183 | 0.354 | 0.419 |
| SSD300[14] | VGG16 | - | [300,300] | 0.232 | 0.412 | 0.234 | 0.530 | 0.232 | 0.396 |
| SSD512[14] | VGG16 | - | [512,512] | 0.268 | 0.465 | 0.278 | 0.900 | 0.289 | 0.419 |
| Faster R-CNN[9] | ResNet-101-FPN | - | [600, 1000] | 0.362 | 0.591 | 0.390 | 0.182 | 0.390 | 0.482 |
| Mask R-CNN[26] | ResNet-101-FPN | - | [800, -] | 0.382 | 0.603 | 0.417 | 0.201 | 0.411 | 0.502 |
| FCOS[18] | ResNet-101-FPN | - | [800,1333] | **0.410** | 0.607 | 0.441 | 0.240 | 0.441 | 0.510 |
| RetinaNet[6] | ResNet-50-FPN | 1x | [600,1000] | 0.353 | 0.543 | 0.375 | 0.184 | 0.394 | 0.480 |
| RetinaNet | ResNet-101-FPN | 1x | [600,1000] | 0.370 | 0.562 | 0.394 | 0.187 | 0.417 | 0.502 |
| IoU-aware Net[23] | ResNet-50-FPN | 1x | [600,1000] | 0.361 | 0.527 | 0.389 | 0.189 | 0.406 | 0.489 |
| IoU-aware Net | ResNet-101-FPN | 1x | [600,1000] | 0.379 | 0.548 | 0.409 | 0.196 | 0.428 | 0.525 |
| RetinaNet-Conf | ResNet-50-FPN | 1x | [600,1000] | 0.360 | 0.529 | 0.384 | 0.184 | 0.406 | 0.491 |
| RetinaNet-Conf | ResNet-101-FPN | 1x | [600,1000] | 0.380 | 0.552 | 0.407 | 0.191 | 0.431 | 0.524 |
| RetinaNet-Conf | ResNet-101-FPN | 2x | [600,1000] | **0.384** | 0.555 | 0.412 | 0.194 | 0.433 | 0.535 |

Table 6: Comparison of state-of-the-art detectors. There are four parts in this experiment. The first, second and third part display performances of classical one-stage, two-stage detectors and anchor-free detector respectively, and these performances are cited from their proposed literatures. The last part mainly displays the detectors related to our work and the performances are gained by retrained based on MMDetection. Schedule 1x and schedule 2x are same with MMdetection, there is total 12 epochs and there is 0.1 times lr decay on epoch[8, 11] when schedule is 1x, while there is total 24 epochs and there is 0.1 times lr decay on epoch[16,

22] when schedule is 2x.

Table 6 shows performances of state-of-the-art detectors. Compared early works of one-stage detectors, RetinaNet performs great superiority. Improvements of classification loss function and backbone network are the mainly contributions. But there is still nearly 1 point gap between RetinaNet with resnet50 and Faster RCNN with resnet101, yet our work with resnet50 has almost same performance with it. The $AP_{0.50}$ of Faster RCNN is fast more than that of RetinaNet-Conf, it demonstrates prediction results of our work have better localization. Also compared with the baseline RetinaNet, our work can improve performance no matter backbone network is ResNet50 or ResNet101. ResNet50 increases by 0.7 AP and ResNet101 increases 1.0 AP respectively. When using the config of schedule 2x, it can gain 0.384 AP, which is a great improvement on RetinaNet. We can find FCOS can achieve 0.41 AP with ResNet101, but it uses the input size [800, 1333], while the input size of our work is [600, 1000]. As restrict of GPU config, we can't perform the experiment under same config, so it is regret to compare the performance of them under same configs.

## 5 Conclusion

We delve into the misalignment problem of classification and localization in RetinaNet and discover that the poor classification is the main factor of it. So we proposed an object confidence task to improve classification task, also the joint of classification and object confidence is used in inference to take into account the classification and localization confidences in NMS. Without bells and whistles, our work gains +1.0 and +0.7 AP on MS COCO dataset from RetinaNet with resnet101 and resnet50 backbones respectively when there is same training configs. And our work with resnet101 backbone can achieve 0.384 AP on COCO dataset when there is two times training epochs. Besides, the focal loss method eases imbalance of positive and negative samples in one-stage detectors indeed, but there is still plenty of room for classification task improvement. As introduced in this paper, adding extra tasks to improve it is a valid method, because the weight of positive samples in classification head training increases, but designing suitable loss function and training strategy for new tasks is also a critical research subject. The poor classification performance is still a key factor of constraining detection performance, hoping to see more researches on this field in future.